\documentclass[10pt,journal,letterpaper,compsoc]{IEEEtran}

\usepackage{graphics}

\usepackage{amssymb}
\usepackage{amsmath}
\usepackage[pdftex]{graphicx}
\usepackage{multirow}
\usepackage{enumerate}
\usepackage{subfigure}
\usepackage{amsthm}
\theoremstyle{plain}
\newtheorem{thm}{Theorem}[section]

\newtheorem{prop}[thm]{Proposition}

\theoremstyle{definition}
\newtheorem{defn}{Definition}[section]

\theoremstyle{remark}

\begin{document}

\title{Discrete Elastic Inner Vector Spaces with Application to Time Series and Sequence Mining}

\author{Pierre-Francois~Marteau,~\IEEEmembership{Member,~IEEE,}
Nicolas~Bonnel,~
and~Gildas~M\'enier,~

\IEEEcompsocitemizethanks{\IEEEcompsocthanksitem P.-F. Marteau is with Universit\'e de Bretagne Sud, IRISA (UMR 6074), Campus de Tohannic, 56000 Vannes, France \protect\\
E-mail: pierre-francois.marteau AT univ-ubs.fr}

\IEEEcompsocitemizethanks{\IEEEcompsocthanksitem N. Bonnel is with Universit\'e de Bretagne Sud, IRISA (UMR 6074), Campus de Tohannic, 56000 Vannes, France \protect\\
E-mail: nicolas.bonnel AT univ-ubs.fr}

\IEEEcompsocitemizethanks{\IEEEcompsocthanksitem G. M\'enier is with Universit\'e de Bretagne Sud, IRISA (UMR 6074), Campus de Tohannic, 56000 Vannes, France \protect\\
E-mail: gildas.menier AT univ-ubs.fr}

\thanks{}}

\markboth{Journal of \LaTeX\ Class Files, March ~2012}%
{P.-F. Marteau, IEEEtran.cls for Computer Society Journals}

\IEEEcompsoctitleabstractindextext{%
\begin{abstract}
This paper proposes a framework  dedicated to the construction of what we call \textit{discrete elastic inner product}  allowing one to embed sets of non-uniformly sampled multivariate \textit{time} series or sequences of varying lengths into inner product space structures. This framework is based on a recursive definition that covers the case of multiple embedded \textit{time elastic} dimensions. We prove that such inner products exist in our general framework and show how a simple instance of this inner product class operates on some prospective applications, while generalizing the Euclidean inner product. Classification experimentations on time series and symbolic sequences datasets demonstrate the benefits that we can expect by embedding time series or sequences into elastic inner spaces rather than into classical Euclidean spaces. These experiments show good accuracy when compared to the euclidean distance or even dynamic programming algorithms while maintaining a linear algorithmic complexity at exploitation stage, although a quadratic indexing phase beforehand is required.
\end{abstract}

\begin{IEEEkeywords}
Vector space, Discrete time series, Sequence mining, Non-uniform sampling, Elastic inner product, Time warping.
\end{IEEEkeywords}}

\maketitle

\IEEEdisplaynotcompsoctitleabstractindextext

\IEEEpeerreviewmaketitle

\section{Introduction}
\label{Intro}
\IEEEPARstart{T}{ime} series analysis in metric spaces has attracted much attention over numerous decades and in various domains such as biology, statistics, sociology, networking, signal processing, etc, essentially due to the ubiquitous nature of time series, whether they are symbolic or numeric. Among other characterizing tools, time warp distances (see \cite{TWED:Velichko70}, \cite{TWED:Sakoe71}, and more recently \cite{TWED:Chen04}, \cite{TWED:Marteau09} among other references) have shown some interesting robustness compared to the Euclidean  metric especially when similarity searching in time series data bases is an issue. Unfortunately, this kind of elastic distance does not enable direct construction of definite kernels which are useful when addressing regression, classification or clustering of time series. A fortiori, they do not make it possible to directly construct inner products involving some \textit{time elasticity}, which are basically able to cope with some stretching or some compression along specific dimension.  Recently, \cite{TWED:Marteau2010} have shown that it is quite easy to propose inner product with time elasticity capability at least for some restricted time series spaces, basically spaces containing uniformly sampled time series, all of which have the same lengths (in such cases, time series can be embedded easily in Euclidean spaces).\\

The aim of this paper is to derive an extension from this preliminary work for the construction of time elastic inner products, to achieve the construction of an elastic inner product structure for a \textit{quasi-unrestricted} set of sequential data or time series, i.e. sets for which the data is not necessarily uniformly sampled and may have any lengths. Section two of the paper gives the main notations used throughout this paper and presents a recursive construction for inner-like products. It then gives the conditions and the proof of existence of  elastic inner products (and elastic vector spaces) defined on a \textit{quasi-unrestricted} set of time series while explaining what we mean by \textit{quasi-unrestricted}. The third section succinctly presents some  preliminary applications, mainly to highlight some of the features of elastic inner product vector spaces such as orthogonality. The fourth section presents two effective experimentations, the first one relating to time series classification and the second one addressing symbolic sequences classification.   \\

\section{Elastic Inner Product Vector Spaces}
\label{EVS}
Starting from the recursive structure of the Dynamic Time Warping (DTW) equation in which the non linear $Max$ operator is replaced by a $Sum$ operator, we propose a quite general and parameterized recursive equation that we call \textit{elastic product}. Our goal in this section is to derive the conditions for which such \textit{elastic product} is an \textit{inner product} that maintains a form of \textit{time elasticity}.

\subsection{Sequence and sequence element}

\begin{defn}
\label{Sequence and sequence element}
Given a finite sequence $A$ we denote by $A(i)$ the $i^{th}$ element (symbol or sample) of sequence $A$. We will consider that $A(i) \in S \times T$ where $(S,\oplus_S, \otimes_S)$ is a vector space that embeds the multidimensional \textit{space} variables (e.g. $S \subset \mathbb{R}^d$, with $d \in \mathbb{N}^+$) and $T \subset \mathbb{R}$ embeds the \textit{timestamps} variable, so that we can write $A(i)=(a(i),t_{a(i)})$ where $a(i) \in S$ and $t_{a(i)} \in T$, with the condition that $t_{a(i)}> t_{a(j)}$ whenever $i>j$ (timestamps strictly increase in the sequence of samples).
$A_{i}^{j}$ with $i \leq j$ is the subsequence consisting of the $i_{th}$ through the $j_{th}$ element (inclusive) of $A$. So $A_{i}^{j}=A(i)A(i+1)...A(j)$. $\Lambda$ denotes the null element. By convention $A_{i}^{j}$ with $i>j$ is the null time series, e.g. $\Omega$.\\
\end{defn}

\subsection{Sequence set}
\begin{defn}
\label{Sequence set}
The set of all finite discrete time series is thus embedded in a spacetime characterized by a single discrete \textit{temporal} dimension, that encodes the timestamps, and any number of \textit{spatial} dimensions that encode the value of the time series at a given timestamp.
 We note $\mathbb{U}$ $= \{A_{1}^{p} | p\in \mathbb{N}\}$ the set of all finite discrete time series. $A_{1}^{p}$ is a time series with discrete index varying between $1$ and $p$. We note $\Omega$ the empty sequence (with null length) and by convention $A_{1}^{0}=\Omega$ so that $\Omega$ is a member of set $\mathbb{U}$. $|A|$ denotes the length of the sequence $A$.
Let $\mathbb{U}_p$ = $\{A \in \mathbb{U}\ | \  |A|\ \le p \}$ be the set of sequences whose length is shorter or equal to $p$. Finally let $\mathbb{U}^*$ be the set of discrete time series defined on $(S-\{0_S\}) \times T$, i.e. the set of time series that do not contain the null \textit{spatial} value. We denote by $0_S$ the null value in $S$.\\
\end{defn}

\subsection{Scalar multiplication on $\mathbb{U}^*$}
\begin{defn}
\label{Def_Otimes}
For all $A \in \mathbb{U}^*$ and all $\lambda \in \mathbb{R}$, $C = \lambda \otimes A $ $\in \mathbb{U}^*$ is such that 
for all $i \in \mathbb{N}$ such that $0\le i \le |A|$, $C(i)=(\lambda.a(i), t_{a(i)})$ and thus $|C|=|A|.$\\
\end{defn}

\subsection{Addition on $\mathbb{U}^*$}
\begin{defn}
\label{Def_Oplus}
For all $(A, B) \in (\mathbb{U^*})^2$, the addition of $A$ and $B$, noted $C = A \oplus B $ $\in \mathbb{U}^*$, is defined in a constructive manner as follows:
let $i,j$ and $k$ be in $\mathbb{N}$.
\begin{enumerate}[]
 \item $k=i=j=1$,
 \item As long as $1 \le i \le |A|$ and $1 \le j \le |B|$,
 \begin{enumerate}[]
 \item if $t_{a(i)} < t_{b(j)}$, $C(k)=(a(i),t_{a(i)})$ and $i \leftarrow i+1, k \leftarrow k+1$
 \item else if $t_{a(i)} > t_{b(j)}$, $C(k)=(b(j),t_{b(j)})$ and $j \leftarrow j+1, k \leftarrow k+1$ 
 \item else if $a(i) + b(j) \neq 0$, $C(k)=(a(i)+b(j),t_{a(i)})$ and $i \leftarrow i+1, j \leftarrow j+1, k \leftarrow k+1$ 
 \item else $i \leftarrow i+1, j \leftarrow j+1$\\
\end{enumerate}
\end{enumerate}
\end{defn}

\begin{figure*}[ht!]
\centering
\includegraphics[scale=0.75]{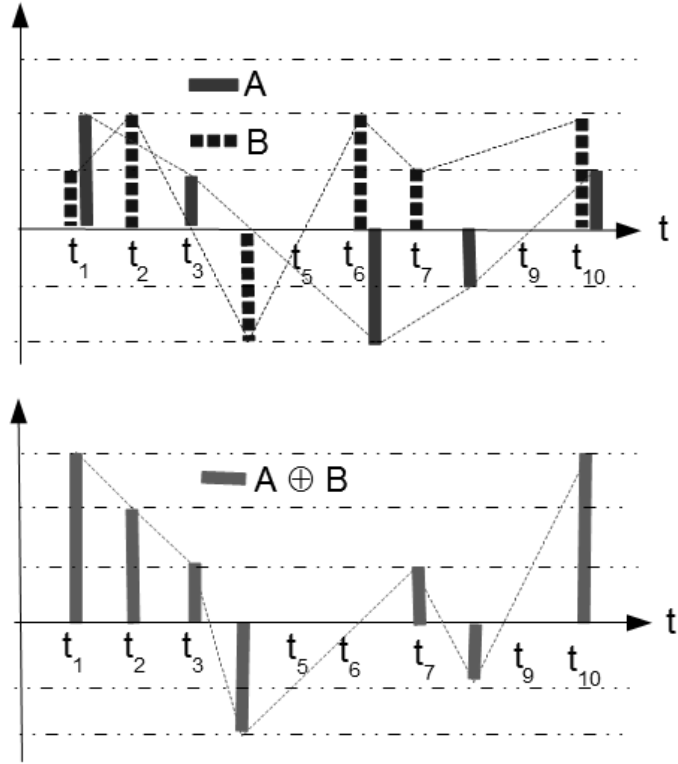}
\caption{The $\oplus$ binary operator when applied to two discrete time series of variable lengths and not uniformly sampled. Co-occurring events have been slightly separated at the top of the figure for readability purposes.}

\label{fig:AplusB}
\end{figure*}

Three comments need to be made at this level to clarify the semantic of the operator $\oplus$:
\begin{enumerate}[i)]
\item Note that the $\oplus$ addition of two time series of equal lengths and uniformly sampled coincides with the classical addition in vector spaces. Fig. \ref{fig:AplusB} gives an example of the addition of two time series that are not uniformly sampled and that have different lengths, except that zero values are discarded to ensure that the sum of two time series will remain a member of $\mathbb{U^*}$.
\item Implicitly (in light of the last case described in Def. \ref{Def_Oplus}), any sequence element of the sort $(0_S, t)$, where $0_S$ is the null value in $S$ and $t \in T$ must be assimilated to the null sequence element $\Lambda$. For instance, the addition of $A=(1,1)(1,2)$ with $B=(-1,1)(1,2)$ is $C= A\oplus B = (2,2)$: the addition of the two first sequence elements is $(0,1)$ that is assimilated to $\Lambda$ and as such suppressed in $C$.
\item The $\oplus$ operator, when restricted to the set $\mathbb{U}^*$ is reversible in that if $C= A \oplus B$ then $A = C \oplus ((-1)\otimes B)$ or $B = C \oplus ((-1) \otimes A)$. This is not the case if we consider the entire set $\mathbb{U}$.
\end{enumerate}

\subsection{Elastic product ($ep$)}
\begin{defn}
\label{ep}
A function $<.,.>: \mathbb{U}^* \times \mathbb{U}^* \rightarrow \mathbb{R}$ is called an Elastic Product if, there exists a function $f: S^2 \rightarrow \mathbb{R}$, a strictly positive function $g: T^2 \rightarrow \mathbb{R}^+$ and three constants $\alpha$, $\beta$ and $\xi$ in $\mathbb{R}$ such that, for any pair of sequences $A_{1}^{p},B_{1}^{q}$,  the following recursive equation holds:

\begin{eqnarray}
\label{Eq.4}
\begin{array}{ll}
 &<A_{1}^{p},B_{1}^{q}>_{ep}=  \\
 & \sum \left\{
   \begin{array}{ll}
     \alpha\cdot<A_{1}^{p-1},B_{1}^{q}>_{ep}  \\
     \beta\cdot<A_{1}^{p-1},B_{1}^{q-1}>_{ep} + f(a(p),b(q))\cdot g(t_{a(p)},t_{b(q)}) \\
     \alpha\cdot<A_{1}^{p},B_{1}^{q-1}>_{ep}  \\
   \end{array}
   \right.
  \end{array}
\end{eqnarray}
\end{defn}

This recursive definition requires defining an initialization. To that end we set, $\forall A \in \mathbb{U}^*$, $<A,\Omega>_{ep}=<\Omega,A>_{ep}=<\Omega,\Omega>_{ep}=\xi$, where $\xi$ is a real constant (typically we set $\xi=0$), and $\Omega$ is the null sequence, with the convention that ${A_i}^j=\Omega$ whenever $i>j$.  \\

This paper addresses the most interesting question of the existence of elastic inner products on the set $\mathbb{U}^*$, i.e. without any restriction on the lengths of the considered time series nor the way they are sampled. If the choice of functions $f$ and $g$, although constrained, is potentially large, we show hereinafter that the choice for constants $\alpha$, $\beta$ and $\xi$ is unique. 

\subsection{Existence of elastic inner products($eip$) defined on $\mathbb{U}^*$}

\begin{thm}
\label{Main theorem}
$<.,.>_{ep}$ is an inner product on $(\mathbb{U}^*, \oplus, \otimes)$ iff:\\
\begin{enumerate}[i)]
\item $\xi = 0$.
\item $g:(T \times T) \rightarrow \mathbb{R}$ is symmetric and strictly positive,
\item $f$ is an inner product on $(S,\oplus_S, \otimes_S)$, if we extend the domain of $f$ on $S$ while setting $f(0_S,0_S)=0$. 
\item $\alpha=1$ and $\beta=-1$,
\end{enumerate} 
\end{thm}

\subsubsection{proof of theorem \ref{Main theorem}}

\textbf{Proof of the direct implication}\\
Let us suppose first that $<.,.>_{ep}$ is an inner product defined on $\mathbb{U}^*$. Then since $<.,.>_{ep}$ is positive-definite necessarily $<\Omega, \Omega>_{ep}=\xi=0$. Furthermore, for any $A_1=(a,t_1)$ and $A_2=(a,t_2) \in \mathbb{U}^*$ with $a\neq 0_S$, $<A_1,A_2>_{ep}=g(t_1,t_2).f(a,a) \neq 0$. Since $<.;.>_{ep}$ is symmetric, we get that $g(t_1,t_2)=g(t_2,t_1)$ for any $(t_1,t_2) \in T^2$ which establishes that $g$ is symmetric. Since $g$ is strictly positive by definition of $<.,.>_{ep}$, i) and ii) are satisfied.\\

For any $A=(a,t_a) \in \mathbb{U}^*$, $<A,A>_{ep}=f(a,a)g(t_a,t_a)>0$. Since $g$ is strictly positive, then we get that $f(a,a)>0$. If we set $f(0_S,0_S)=0$, we establish that $f$ is positive-definite on $S$.\\

Since $\xi=0$, for any $A$, $B$, and $C \in \mathbb{U}^*$ such that $A=(a, t)$, $B(b,t)$ and $C=(c,t_c)$, we have:\\
$<A \oplus B,C>_{ep}=f(a \oplus_S b, c).g(t,t_c)$. \\
As $<A \oplus B,C>_{ep}=<A, C>_{ep}+<B,C>_{ep}$ \\
$=f(a,c).g(t,t_c)+f(b,c).g(t,t_c)=(f(a,c)+f(b,c)).g(t,t_c)$,\\
As $g$ is strictly positive, we get that $f(a \oplus_S b, c)=(f(a,c)+f(b,c))$.\\
Furthermore, $<\lambda \otimes A,C>_{ep}=f(\lambda \otimes_S a, c).g(t,t_c)$. \\
As $<\lambda \otimes A,C>_{ep}=\lambda.<A, C>_{ep}=\lambda.f(a,c).g(t,t_c)$ and $g$ is strictly positive, we get that $f(\lambda \otimes_S a, c)= \lambda.f(a,c)$.\\
This shows that $f$ is linear, symmetric and positive-definite. Hence it is an inner product on $(S,\oplus_S, \otimes_S)$ and iii) is satisfied.\\

Let us show that necessarily $\alpha=1$ and $\beta=-1$. To that end, let us consider any $A_{1}^p, B_{1}^q$ and $C_{1}^r$ in $\mathbb{U^*}$, such that $p>1, q>1, r>1$ and such that $t_{a(p)}<t_{b(q)}$, i.e.  if $X_{1}^{s}=A_{1}^{p} \oplus B_{1}^{q}$, then $X_{1}^{s-1}=A_{1}^{p} \oplus B_{1}^{q-1}$.\\
Since by hypothesis $<.,.>_{ep}$ is an inner product $(\mathbb{U}^*, \oplus, \otimes)$, it is linear and thus we can write:\\
$<A_{1}^p \oplus B_{1}^q,C_{1}^r>_{ep}=<A_{1}^p,C_{1}^r>_{ep}+<B_{1}^q,C_{1}^r>_{ep}$.\\

Decomposing $<A_{1}^p \oplus B_{1}^q,C_{1}^r>_{ep}$, we obtain:\\
$<A_{1}^p \oplus B_{1}^q,C_{1}^r>_{ep}=\alpha.<A_{1}^{p} \oplus B_{1}^{q-1}, C_{1}^r>_{ep} + $\\
$\beta.<A_{1}^{p} \oplus B_{1}^{q-1}, C_{1}^{r-1}>_{ep} + f(b(q),c(r)).g(t_{b(q)}, t_{c(r)}) + \alpha.<A_{1}^p \oplus B_{1}^q,C_{1}^{r-1}>_{ep}$\\
As $<.,.>_{ep}$ is linear we get:\\
$<A_{1}^p \oplus B_{1}^q,C_{1}^r>_{ep}=\alpha.<A_{1}^{p},C_{1}^r>_{ep} + \alpha.<B_{1}^{q-1}, C_{1}^r>_{ep} + $\\
$\beta.<A_{1}^{p}, C_{1}^{r-1}>_{ep} + \beta.<B_{1}^{q-1}, C_{1}^{r-1}>_{ep} + f(b(q),c(r)).g(t_{b(q)}, t_{c(r)}) + $\\
$\alpha.<A_{1}^p,C_{1}^{r-1}>_{ep} + \alpha.<B_{1}^q,C_{1}^{r-1}>_{ep}$\\
Hence,\\
$<A_{1}^p \oplus B_{1}^q,C_{1}^r>_{ep}=\alpha.<A_{1}^{p},C_{1}^r>_{ep}+\beta.<A_{1}^{p},C_{1}^{r-1}>_{ep} +$\\
$\alpha.<A_{1}^{p},C_{1}^{r-1}>_{ep} + <B_{1}^{q}, C_{1}^{r}>_{ep}$\\

If we decompose $<A_{1}^{p},C_{1}^r>_{ep}$, we get:\\
$<A_{1}^p \oplus B_{1}^q,C_{1}^r>_{ep}=(\alpha^2+\beta+\alpha)<A_{1}^p,C_{1}^{r-1}>_{ep}+\alpha.\beta.<A_{1}^{p-1},C_{1}^{r-1}>_{ep}+\alpha.f(a(p),c(r)).g(t_{a(p)},t_{c(r)})+ \alpha^2.<A_{1}^{p-1},C_{1}^{r}>_{ep} + <B_{1}^q,C_{1}^{r}>_{ep}$\\

Thus we have to identify $<A_{1}^p,C_{1}^{r}>_{ep}=\alpha.<A_{1}^p,C_{1}^{r-1}>_{ep} + \beta.<A_{1}^{p-1},C_{1}^{r-1}>_{ep} + f(a(p),c(r)).g(t_{a(p)},t_{c(r)}) + \alpha.<A_{1}^{p-1},C_{1}^{r}>_{ep}$\\
with $(\alpha^2+\beta+\alpha)<A_{1}^p,C_{1}^{r-1}>_{ep}+\alpha.\beta.<A_{1}^{p-1},C_{1}^{r-1}>_{ep}+\alpha.f(a(p),c(r)).g(t_{a(p)},t_{c(r)})+ \alpha^2.<A_{1}^{p-1},C_{1}^{r}>_{ep}$.\\

The unique solution is $\alpha=1$ and $\beta=-1$. That is if $<.,.>_{ep}$ is an existing inner product, then necessarily $\alpha=1$ and $\beta=-1$, establishing iv).\\

\textbf{Proof of the converse implication}\\
Let us suppose that i), ii), iii) and iv) are satisfied and show that $<.,.>_{ep}$ is an inner product on $\mathbb{U}^*$.\\

First, by construction, since $f$ and $g$ are symmetric, so is $<.,.>_{ep}$. \\

It is easy to show by induction that $<.,.>_{ep}$ is non-decreasing with the length of its arguments, namely, $\forall A_{1}^p$ and $B_{1}^q$ in $\mathbb{U^*}$,\\
$<A_{1}^{p}, B_{1}^{q}>_{ep}-<A_{1}^{p}, B_{1}^{q-1}>_{ep} \ge 0$. Let $n=p+q$. The proposition is true at rank $n=0$. It is also true if $A_{1}^p=\Omega$, whatever $B_{1}^q$ is, or $B_{1}^q=\Omega$, whatever $<A_{1}^{p}$ is. Suppose it is true at a rank $n \ge 0$, and consider $A_{1}^p \neq \Omega$ and $B_{1}^q \neq \Omega$ such that $p+q=n$.\\
By decomposing $<A_{1}^{p}, B_{1}^{q}>_{ep}$ we get:\\
$<A_{1}^{p}, B_{1}^{q}>_{ep}-<A_{1}^{p}, B_{1}^{q-1}>_{ep} = - <A_{1}^{p-1}, B_{1}^{q-1}>_{ep} + f(a(p),b(q)).g(t_{a(p)}, t_{b(q)}) + <A_{1}^{p-1}, B_{1}^{q}>_{ep}$\\
Since $f(a(p),b(q)).g(t_{a(p)}, t_{b(q)}) > 0$ and the proposition is true by inductive hypothesis at rank $n$, we get that $<A_{1}^{p}, B_{1}^{q}>_{ep}-<A_{1}^{p}, B_{1}^{q-1}>_{ep}) > 0$. By induction the proposition is proved.\\

Let us show by induction on the length of the time series the positive definiteness of $<.,.>_{ep}$.\\
At rank $0$ we have $<\Omega,\Omega>_{ep}=\xi=0$. At rank $1$, let us consider any time series of length $1$, $A_{1}^1$. $<A_{1}^1,A_{1}^1>_{ep}=f(a(1),a(1)).g(t_{a(1)},t_{a(1)})>0$ by hypothesis on $f$ and $g$. Let us suppose that the proposition is true at rank $n>1$ and let consider
 any time series of length $n+1$, $A_{1}^{n+1}$. Then, since $\alpha=1$ and $\beta=-1$, we get\\
$<A_{1}^{n+1},A_{1}^{n+1}>_{ep}=2.<A_{1}^{n+1},A_{1}^{n}>_{ep} - <A_{1}^{n},A_{1}^{n}>_{ep} \\ \hspace*{0.3in}+ f(a(n+1),a(n+1)).g(t_{a(n+1)},t_{a(n+1)})$. \\
Since $<A_{1}^{n+1},A_{1}^{n}>_{ep} - <A_{1}^{n},A_{1}^{n}>_{ep} \ge 0$, and $f(a(n+1),a(n+1)).g(t_{a(n+1)},t_{a(n+1)})>0$, $<A_{1}^{n+1},A_{1}^{n+1}>_{ep}>0$, showing that the proposition is true at rank $n+1$. By induction, the  proposition is proved, which establishes the positive-definiteness of $<.,.>_{ep}$ since $<A_{1}^{p}, A_{1}^{p}>_{ep}=0$ only if $A_{1}^{p}=\Omega$.\\

Let us consider any $\lambda \in \mathbb{R}$, and any $A_{1}^p, B_{1}^q$ in $\mathbb{U^*}$ and show by induction on $n=p+q$ that$<\lambda \otimes A_{1}^p, B_{1}^q>_{ep}=\lambda.<A_{1}^p, B_{1}^q>_{ep}$: \\
The proposition is true at rank $n=0$. Let us suppose that the proposition is true at rank $n \ge 0$, i.e. for all $r\le n$, and consider any pair $A_{1}^p, B_{1}^q$ of time series such that $p+q=n+1$.\\
We have: $<\lambda \otimes A_{1}^p, B_{1}^q>_{ep}=\alpha.<\lambda \otimes A_{1}^p, B_{1}^{q-1}>_{ep}+\beta.<\lambda \otimes A_{1}^{p-1}, B_{1}^{q-1}>_{ep}+f(\lambda\otimes_S a(p), b(q)).g(t_{a(p)},t_{b(q)}) + \alpha.<\lambda \otimes A_{1}^{p-1}, B_{1}^{q}>_{ep}$\\
Since $f$ is linear on $(S, \oplus_S, \otimes_S)$, and since the proposition is true by hypothesis at rank $n$, we get that $<\lambda \otimes A_{1}^p, B_{1}^q>_{ep}=\lambda.\alpha <A_{1}^p, B_{1}^{q-1}>_{ep} + \lambda.\beta.<A_{1}^{p-1}, B_{1}^{q-1}>_{ep}+\lambda.f(a(p), b(q)).g(t_{a(p)},t_{b(q)}) + \lambda.\alpha.<A_{1}^{p-1}, B_{1}^{q}>_{ep}=\lambda.<A_{1}^p, B_{1}^q>_{ep}$.\\
By induction, the proposition is true for any $n$, and we have proved this proposition.

Furthermore, for any $A_{1}^p, B_{1}^q$ and $C_{1}^r$ in $\mathbb{U^*}$, let us show by induction on $n=p+q+r$ that $< A_{1}^p \oplus B_{1}^q, C_{1}^r>_{ep}=< A_{1}^p, C_{1}^r>_{ep}+< B_{1}^q, C_{1}^r>_{ep}$. Let $X_{1}^s$ be equal to $A_{1}^p \oplus B_{1}^q$. The proposition is obviously true at rank $n=0$. Let us suppose that it is true up to rank $n \ge 0$, and consider any $A_{1}^p, B_{1}^q$ and $C_{1}^r$ such that $p+q+r=n+1$. 

Three cases need then to be considered:
\begin{enumerate}[1)]
\item if $X_{1}^{s-1}=A_{1}^{p-1} \oplus B_{1}^{q-1}$, then $t_{a(p)}=t_{b(q)}=t$ and \\
$<A_{1}^p \oplus B_{1}^q, C_{1}^r>_{ep}=\alpha.< A_{1}^p \oplus B_{1}^q, C_{1}^{r-1}>_{ep} +$\\  \hspace*{10mm}$\beta.<A_{1}^{p-1} \oplus B_{1}^{q-1}, C_{1}^{r-1}>_{ep} +$ \\
\hspace*{10mm}$f((a(p)+b(q)), c(r)).g(t,t_{c(r)})+$ \\
\hspace*{10mm}$\alpha.<A_{1}^{p-1} \oplus B_{1}^{q-1}, C_{1}^{r}>_{ep}$. \\
Since $f$ is linear on $(S, \oplus_S, \otimes_S)$, and the proposition true at rank $n$, we get the result.
\item if $X_{1}^{s-1}=A_{1}^{p} \oplus B_{1}^{q-1}$, then $t_{a(p)} < t_{b(q)}=t$ and \\
$<A_{1}^p \oplus B_{1}^q, C_{1}^r>_{ep}=\alpha.< A_{1}^p \oplus B_{1}^q, C_{1}^{r-1}>_{ep} + \beta.<A_{1}^{p} \oplus B_{1}^{q-1}, C_{1}^{r-1}>_{ep} +f(b(q), c(r)).g(t,t_{c(r)})+ \alpha.<A_{1}^{p} \oplus B_{1}^{q-1}, C_{1}^{r}>_{ep}$. Having $\alpha=1$ and $\beta=-1$ with the proposition supposed to be true at rank $n$ we get the result.
\item if $X_{1}^{s-1}=A_{1}^{p-1} \oplus B_{1}^{q-1}$, we proceed similarly to case 2).
\end{enumerate}

Thus the proposition is true at rank $n+1$, and by induction the proposition is true for all $n$. This establishes the linearity of $<.,.>_{ep}$.\\
This ends the proof of the converse implication and theorem \ref{Main theorem} is therefore established $\square$\\.

The existence of functions $f$ and $g$ entering into the definition of $<.,.>_{ep}$ and satisfying the conditions allowing for the construction of an  inner product on $(\mathbb{U}^*, \oplus, \otimes)$ is ensured by the following proposition:
\begin{prop}
\label{PropExistence}
The functions $f:S^2 \rightarrow \mathbb{R}$ defined as $f(a,b) = <a, b>_S$ where $<.,.>_S$ is an inner product on $(S, \oplus_S, \otimes_S)$ and $g:T^2 \rightarrow \mathbb{R}$ defined as $f(t_a,t_b)=e^{-\nu \cdot d(t_a,t_b)}$, where $d$ is a distance defined on $T^2$ and $\nu \in \mathbb{R}^+$, satisfy the conditions required to construct an elastic inner product on $(\mathbb{U^*}, \oplus, \otimes)$.
\end{prop}

The proof of Prop.\ref{PropExistence} is obvious. This proposition establishes the existence of $ep$ inner products, that we will denote $eip$ (Time Elastic Inner Product). An eip as thus the following structure:

\begin{eqnarray}
\label{Eq.eip}
\begin{array}{ll}
 &<A_{1}^{p},B_{1}^{q}>_{eip}=  \\
 & \sum \left\{
   \begin{array}{ll}
     <A_{1}^{p-1},B_{1}^{q}>_{eip}  \\
     -<A_{1}^{p-1},B_{1}^{q-1}>_{eip} + \\
     \hspace{.25in} g(t_{a(p)},t_{b(q)})\cdot<(a(p),b(q)>_{eip(S)} \\
     <A_{1}^{p},B_{1}^{q-1}>_{eip}  \\
   \end{array}
   \right.
  \end{array}
\end{eqnarray}

With the initialization: $\forall A \in \mathbb{U}^*$, $<A,\Omega>_{eip}=<\Omega,A>_{eip}=<\Omega,\Omega>_{eip}=\xi$, where $\xi$ is a real constant (typically we set $\xi=0$), and $\Omega$ is the null sequence, with the convention that ${A_i}^j=\Omega$ whenever $i>j$.  \\

Note that $<.,.>_S$ can be chosen to be a $eip$ as well, in the case where a second \textit{time elastic} dimension is required. This leads naturally to recursive definitions for $ep$ and $eip$.\\

\begin{prop}
\label{PropEuclProdvsEIP}
For any $n \in \mathbb{N}$, and any discrete subset $T=\{t_1, t_2, \cdots, t_n\} \subset \mathbb{R}$, let $\mathbb{U}_{n,\mathbb{R}, T}$ be the set of all time series defined on $\mathbb{R} \times T$ whose lengths are $n$ (the time series in $\mathbb{U}_{n,\mathbb{R}, T}$ are considered to be uniformly sampled). Then, the $eip$ on $\mathbb{U}_{n,\mathbb{R}}$ constructed from the functions $f$ and $g$ defined in Prop. \ref{PropExistence} tends towards the Euclidean inner product when $\nu \rightarrow \infty$ if $S$ is an Euclidean space and $<a,b>_S$ is the Euclidean inner product defined on $S$.
\end{prop}

The proof of Prop.\ref{PropEuclProdvsEIP} is straightforward and is omitted. This proposition shows that $eip$ generalizes the classical Euclidean inner product.\\

\subsection{Algorithmic complexity}
\label{algorithmic complexity}

The general complexity associated to the calculation of the $eip$ of two time series of lengths $p$ and $q$ as specified by Eq.\ref{Eq.eip} is $O(p \cdot q)$. Basically this is the same complexity as that required for the calculation of the complete dynamic programming solutions such as the Dynamic Time Warping (DTW) \cite{TWED:Velichko70} \cite{TWED:Sakoe71} algorithm evaluated on these two time series. Nevertheless, as discussed in section \ref{IR}, Prop. \ref{PropRnEmbedding} allows for efficient implementations of retrieval process (in linear time complexity) once a straightforward indexing phase has been implemented. This result is demonstrated in practice the experimentation section (sec. \ref{Expe}).

\section{Some preliminary applications}

We present in the following sections some applications to highlight the properties of Elastic Inner Product Vector Spaces ($EIPVS$).

\subsection{Distance in $EIPVS$}

The following proposition provides $\mathbb{U}^*$ with a norm and a distance, both induced by a $eip$.

\begin{prop}
\label{eip_distance}
For all $A_{1}^{p} \in \mathbb{U}^*$, and any $<.,.>_{eip}$  defined on $(\mathbb{U}^*, \oplus, \otimes)$  $\sqrt{<A_{1}^{p},A_{1}^{p}>_{eip}}$ is a norm on $\mathbb{U}^*$.\\
For all pair $(A_{1}^{p},B_{1}^{q}) \in (\mathbb{U}^*)^2$, and any $eip$  defined on $(\mathbb{U}^*, \oplus, \otimes)$,  \\
$\delta_{eip}(A_{1}^{p},B_{1}^{q})= \sqrt{<A_{1}^{p} \oplus (-1.\otimes B_{1}^{q}), A_{1}^{p} \oplus (-1.\otimes B_{1}^{q})>_{eip}}$\\
$ = \sqrt{<A_{1}^{p},A_{1}^{p}>_{eip}+<B_{1}^{q},B_{1}^{q}>_{eip} -2\cdot<A_{1}^{p},B_{1}^{q}>_{eip}}$ \\
defines a distance metric on $\mathbb{U}^*$.
\end{prop}

The proof of Prop. \ref{eip_distance} is straightforward and is omitted. \\

\subsection{Orthogonalization in $EIPVS$}

To exemplify the effect of elasticity in $EIPVS$, we give below the result of the Gram-Schmidt orthogonalization algorithm for two families of independent univariate time series. The first family is composed of uniformly sampled time series having increasing lengths. The second family (a sine-cosine basis) is composed of uniformly sampled time series, all of which have the same length.  \\

The tests which are described in the next sections were performed on a set $\mathbb{U}^*$ of discrete time series whose elements are defined on $(\mathbb{R}-\{0\} \times [0;1])^2$ using the following $eip$:\\

\begin{eqnarray}
\label{Eq_eip}
\begin{array}{ll}
 &<A_{1}^p, B_{1}^q>_{eip} =  \\
 & \sum \left\{
   \begin{array}{ll}
     <A_{1}^{p},B_{1}^{q-1}>_{eip}  \\
     - <A_{1}^{p-1},B_{1}^{q-1}>_{eip} + \\
     \hspace{1in} a(p)b(q)\cdot e^{-\nu.|t_{a(p)}-t_{b(q)}|^{2}} \\
     <A_{1}^{p-1},B_{1}^{q}>_{eip}  \\
   \end{array}
   \right.
  \end{array}
\end{eqnarray}
 
\begin{figure}[h!]
\centering
\includegraphics[scale=0.25]{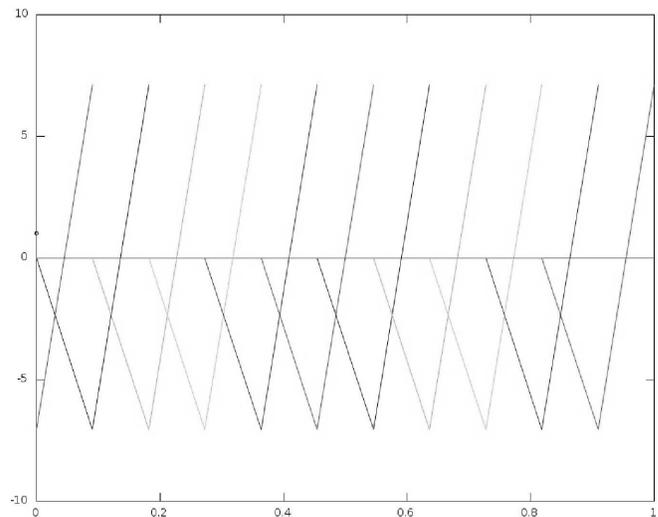}
\caption{Result of the orthogonalization of the family of length time series defined in Eq.\ref{BasisOfIncreasingLength} using $\nu=.01$: except for the first \textit{spike} located at \textit{time} $0$, each original \textit{spike} is replaced by two \textit{spikes}, one negative the other positive.} 
\label{fig:spk_orthogonal}
\end{figure}

\begin{figure}[h!]
\centering
\includegraphics[scale=0.5]{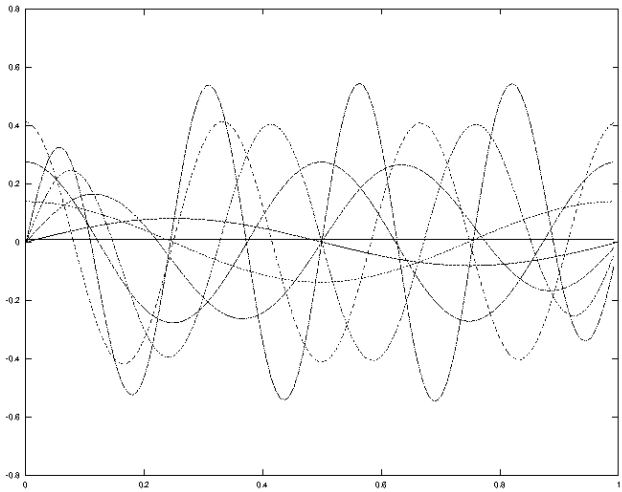}
\caption{Orthogonalization of the \textit{sine-cosine} basis using $\nu=.01$: the waves are slightly deformed jointly in amplitude and in frequency. For readability of the figure, we have presented the 8 first components} 
\label{fig:sin_orthogonal}
\end{figure}

\subsubsection{Orthogonalization of an independent family of time series with increasing lengths}
The family of time series we are considering is composed of $11$ time series uniformly sampled, whose lengths are $11$ samples: 
\begin{eqnarray}
\label{BasisOfIncreasingLength}
\begin{array}{ll}
 (1,0)\\
 (\epsilon,0)(1,1/10)\\
 (\epsilon,0)(\epsilon,0)(1,1/10)\\
 \cdots\\
 (\epsilon,0)(\epsilon,1/10)(\epsilon,2/10) \cdots (1,1)\\
\end{array}
\end{eqnarray}

Since, the zero value cannot be used for the space dimension, we replaced it by $\epsilon$, which is the smallest non zero positive real for our test machine (i.e. $2^{-1074}$). The result of the Gram-Schmidt orthogonalization process using $\nu=.01$ on this basis is given in Fig.\ref{fig:spk_orthogonal}.

\subsubsection{Orthogonalization of a sine-cosine basis}

An orthonormal family of discrete sine-cosine functions is not anymore orthogonal in a $EIPVS$.
The result of the Gram-Schmidt orthogonalization process using $\nu=.01$ when applied on a discrete sine-cosine basis is given in Fig.\ref{fig:sin_orthogonal}, in which only the 8 first components are displayed. The lengths of the waves are 128 samples.

\subsection{Indexing for fast retrieval in time series data bases}
\label{IR}
Prop.\ref{PropEuclProdvsEIP} shows how the elastic inner product generalizes the Euclidean inner product when time series are embedded into a finite dimensional vector space (one dimension per timestamps). We consider in this subsection such embeddings with the convention that if a time series has no value for a given timestamps a zero value is added on the dimension corresponding to the missing timestamps. Each time series is thus represented by a finite dimensional vector, let say in $\mathbb{R}^{n}$. 

\begin{prop}
\label{PropRnEmbedding}

Given a symmetric and strictly positive function $g$, let consider the so-called $n \times n$ symmetric elastic matrix in $\mathbb{R}^{n^{2}}$ defined as:

\[
E =
\left[ {\begin{array}{ccccc}
 g(t_{1},t_{1}) & g(t_{1},t_{2}) & g(t_{1},t_{3}) &  \cdots & g(t_{1},t_{n})\\
 g(t_{2},t_{1}) & g(t_{2},t_{2}) & g(t_{2},t_{3}) &  \cdots & g(t_{2},t_{n}) \\
 g(t_{3},t_{1}) & g(t_{3},t_{2}) & g(t_{3},t_{3}) &  \cdots & g(t_{3},t_{n}) \\
 \cdots &  \cdots & \cdots & \cdots & \cdots  \\
 g(t_{n},t_{1}) & g(t_{n},t_{2}) &  g(t_{n},t_{3}) &  \cdots & g(t_{n},t_{n}) \\
\end{array} } \right]
\]

and two time series $A_{1}^{n}$ and $B_{1}^{n}$ represented by two vectors in $\mathbb{R}^{n}$. Then the elastic inner product defined recursively as: 

\begin{eqnarray}
\label{Eq.eipvsMatrix}
\begin{array}{ll}
 &<A_{1}^{n},B_{1}^{n}>_{eip}=  \\
 & \sum \left\{
   \begin{array}{ll}
     <A_{1}^{n-1},B_{1}^{n}>_{eip}  \\
     -<A_{1}^{n-1},B_{1}^{n-1}>_{eip} + \\
     \hspace{1in} a(n)b(n)\cdot g(t_{a(n)},t_{b(n)}) \\
     <A_{1}^{n},B_{1}^{n-1}>_{eip}  \\
   \end{array}
   \right.
  \end{array}
\end{eqnarray}

identifies to the following matrix products $\left[A_{1}^{n}\right]^{T}E\left[B_{1}^{n}\right]$
\end{prop}

This result is quite interesting in the scope of time series information retrieval especially when addressing very large databases. Let $D=\{B(i)_{1}^{n}\}_{i=1 \cdots m}$ be a time series database of size $m$, and consider the indexing phase that consists in constructing $D_E=\{E \left[ B(i)_{1}^{n} \right]\}_{i=1 \cdots m}$. Then the retrieval of all the time series in $D$ \textit{elastically} similar to a given request $A_{1}^{n}$ will require the computation of $<A_{1}^{n},B(i)_{1}^{n}>_{eip}$, for $i \in \{1,\cdots, m\}$, which reduces to evaluating:

\begin{eqnarray}
\label{Eq.matrixEIP}
\begin{array}{ll}
 <A_{1}^{n},B(i)_{1}^{n}>_{eip}=\left[A_{1}^{n}\right]^{T}\left[E B(i)_{1}^{n}\right]
  \end{array}
\end{eqnarray}
 
 for $i \in \{1,\cdots, m\}$ which is done in linear complexity. Basically this construction breaks the quadratic complexity of the $eip$ at retrieval stage and meets the same linear complexity as the classical Euclidean inner product. 

Moreover one can notice that the $eip$ defined by Eq.\ref{Eq.eipvsMatrix} has it \textit{continuous time} equivalent that expresses as:
$<a,b>=\iint \! a(t)b(\tau)g(t,\tau) \, \mathrm{d}t\mathrm{d}\tau$

\subsection{Kernel methods in $EIPVS$}
A wide range of literature exists on kernel theory, among which \cite{TWED:BergChristensenRessel84}, \cite{TWED:Scholkopf01} and \cite{TWED:Shawe04} present some large syntheses of  major results. We give hereinafter basic definitions and immediate results regarding kernel construction based on $eip$.
\begin{defn}
\label{Kernels}
A kernel on a non empty set $U$ refers to a complex (or real) valued symmetric function $\varphi(x,y) : U \times U \rightarrow \mathbb{C}$ (or $\mathbb{R}$).  
\end{defn}

\begin{defn}
\label{Definite Kernels}
Let $U$ be a non empty set. A function $\varphi: U \times U \rightarrow \mathbb{C}$ is called a positive (resp. negative) definite kernel if and only if it is Hermitian 
(i.e. $\varphi(x,y)=\overline{\varphi(y,x)}$ where the \textit{overline} stands for the conjugate number) for all $x$ and $y$ in $U$ and $\sum_{i,j=1}^n c_i \bar{c_j} \varphi(x_i, x_j) \ge 0$ (resp. $\sum_{i,j=1}^n c_i \bar{c_j} \varphi(x_i, x_j) \le 0$), 
for all $n$ in $\mathbb{N}$, $(x_1,x_2, ..., x_n) \in U^n$ and $(c_1,c_2,...,c_n) \in \mathbb{C}^n$.\\
\end{defn}

\begin{defn}
 \label{Conditionally Definite Kernels}

Let $U$ be a non empty set. A function $\varphi: U \times U \rightarrow \mathbb{C}$ is called a conditionally positive (resp. conditionally negative) definite kernel if and only if it is Hermitian 
(i.e. $\varphi(x,y)=\overline{\varphi(y,x)}$ for all $x$ and $y$ in $U$) and $\sum_{i,j=1}^n c_i \bar{c_j} \varphi(x_i, x_j) \ge 0$ (resp. $\sum_{i,j=1}^n c_i \bar{c_j} \varphi(x_i, x_j) \le 0$),
for all $n \ge 2$ in $\mathbb{N}$, $(x_1,x_2, ..., x_n) \in U^n$ and $(c_1,c_2,...,c_n) \in \mathbb{C}^n$ with $\sum_{i=1}^n c_i=0$. \\
\end{defn}

In the last two above definitions, it is easy to show that it is sufficient to consider mutually different elements in $U$, i.e. collections of distinct elements $x_1, x_2, ..., x_n$. \\

\begin{defn}
 \label{matrix and kernel}
A positive (resp. negative) definite kernel defined on a finite set $U$ is also called a positive (resp. negative) semidefinite matrix. 
Similarly, a positive (resp. negative) conditionally definite kernel defined on a finite set is also called a positive (resp. negative) conditionally semidefinite matrix.
\end{defn}

\subsubsection{Definiteness of $eip$ based kernel}

\begin{prop}
\label{eipKernelDefiniteness}
A $eip$ is a positive definite kernel.\\
\end{prop}

The proof of Prop. \ref{eipKernelDefiniteness} is straightforward and is omitted. The consequence is that numerous definite (positive or negative) kernels can be derived from a $eip$; among other immediate derivations it can be stated that:

\begin{itemize}
\item $(<.,.>_{eip})^{p}$ is positive definite for all $p \in \mathbb{N}$ (polynomial kernel).
\item $\delta_{eip}$ defined by Prop.\ref{eip_distance} is negative definite.
\item $e^{-\nu\cdot\delta_{eip}}$ is positive definite for all $\nu>0$.
\item $e^{-\nu\cdot(\delta_{eip})^{p}}$ is positive definite for all $\nu>0$ and any $0<p \le 2$.
\item $e^{<A,B>_{eip}}$ is positive definite.
\end{itemize}

Some experimentations on Support vector Machine involving \textit{elastic} kernels are reported in Sec.\ref{Expe}.

\subsection{Elastic Cosine similarity in $EIPVS$, with application to symbolic (e.g. textual) information retrieval}
Similarly to the definition of the cosine of two vectors in Euclidean space, we define the elastic cosine of two sequences by using any $ep$ that satisfies the conditions of theorem \ref{Main theorem}.

\begin{defn}
 \label{elastic cosine}
Given two sequences, $A$ and $B$, the \textit{elastic cosine} similarity of these two sequences is given using a time elastic inner product $<X,Y>_{e}$ and the induced norm $\|X\|_e=\sqrt{<X,X>_e}$ as\\
    $\textit{similarity} = eCOS(\theta) = {<A \cdot B>_{e} \over \|A\|_e \|B\|_e}$
\end{defn}

In the case of textual or sequential data information retrieval, namely text matching or sequence matching, the timestamps variable coincides with the index of words into the text or sequence, and the spatial dimensions encode the words or symbol into a given dictionary or alphabet. For instance, in text mining, each word can be represented using a vector whose dimension is the size of the set of concepts (or senses) that covers the conceptual domain associated to the dictionary, and each coordinate value, that is selected into $[0;1]$, encodes the degree of presence of the concept or senses into the considered word. 
In any case, the elastic cosine similarity measure takes value into $[0;1]$, $0$ indicating the lowest possible similarity value between two sequential data and $1$ the greatest possible similarity value between two sequential data. The elastic cosine similarity takes into account the order of occurrence of the words or symbols into the sequential data which could be an advantage compared to the Euclidean cosine measure that does not cope at all with the words or symbols ordering.

Let us consider the following elastic inner product dedicated to text matching. In the following definition, $A_{1}^{p}$ and $B_{1}^{q}$ are sequences of words that represent textual content.
\begin{defn}
\label{defEIP_TM}
\begin{eqnarray}
\begin{array}{ll}
\label{eqEIP_TM}
<A_{1}^{p}, B_{1}^{q}>_{eip_{tm}}=   \\
\hspace*{2mm} \sum \left\{ 
		\begin{array}{ll}
     <A_{1}^{p-1},B_{1}^{q}>_{eip_{tm}} \\
     -<A_{1}^{p-1},B_{1}^{q-1}>_{eip_{tm}} + \\
     \hspace{1in} e^{-\nu.|t_{a(p)}-t_{b(q)}|^{2}}\delta(a(p), b(q))\\
     <A_{1}^{p},B_{1}^{q-1}>_{eip_{tm}} \\
   \end{array}
   \right.
  \end{array}
\end{eqnarray}
where $a(p)$ and $b(q)$ are vectors whose coordinates identify words with weightings, $\delta(a,b)=<a,b>$ is the Euclidan inner product, and $\nu$ a \textit{time stiffness} parameter.\\
\end{defn}

\begin{figure*}[ht!]
\centering
\includegraphics[scale=0.42]{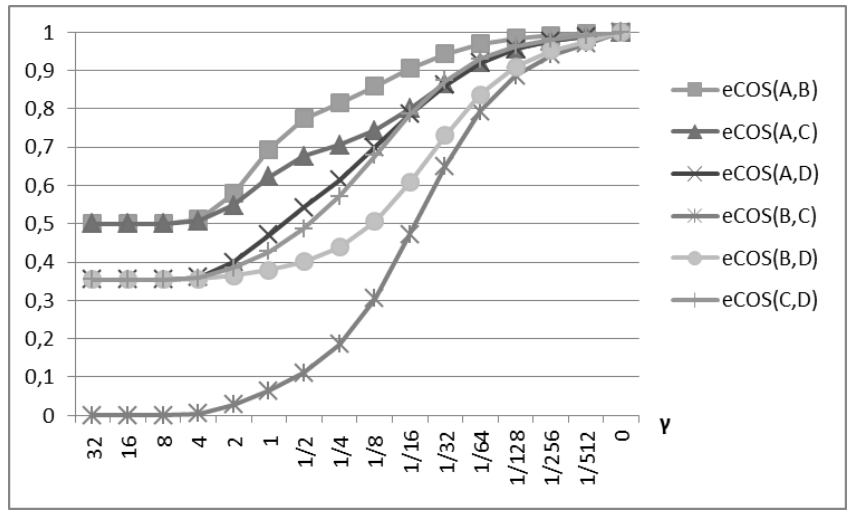}
\caption{Elastic cosine as a function of $\nu$ evaluated on pairwise sequences selected in $\{A=\left[abababab\right],B=\left[aaaabbbb\right],C\left[bbbbaaaa\right],D=\left[bbaa\right]\}$. The timestamps correspond to the index of occurrence of the symbol into the sequence ($i=1,2,\cdots, $).} 
\label{fig:eCOS-toy}
\end{figure*}

\begin{prop}
 \label{Deip_vs_vectorModel}
For $\nu=0$ and $\delta$ redefined as $\delta(a,b)=1$ if $a=b$, $0$ otherwise, the elastic inner product defined in Eq.\ref{eqEIP_TM} coincides with the euclidean inner product between two vectors whose coordinates correspond to term frequencies observed into the $A_{1}^{p}$ and $B_{1}^{q}$ text sequences. If, we change the definition of $\delta$ by the $\delta(a,b)=(IDF(a))^2$ if $a=b$, $0$ otherwise, where $IDF(a)$ is the inverse document frequency of term $a$ into the considered collection, then for $\nu=0$, $<A_{1}^{p}, B_{1}^{q}>_{eip_{tm}}$ coincides with the euclidean inner product between two vectors whose coordinates correspond to the TF-IDF (term frequency times the inverse document frequency) of terms occurring into the $A_{1}^{p}$ and $B_{1}^{q}$ text sequences.
\end{prop}

The proof of proposition \ref{Deip_vs_vectorModel} is straightforward an is omitted.\\

Thus, the elastic cosine measure derived from the elastic inner product defined by Eq.\ref{eqEIP_TM} generalizes somehow the cosine measure implemented in the vector space model \cite{TWED:SaltonG84} and commonly used in the text information retrieval community.

To exemplify the behavior of the elastic cosine similarity on sequential data, we consider the four sequences depicted in Eq.\ref{eq:toy-seq}. The variations of the elastic cosine as a function of $\nu$ evaluated on pairwise sequences are reported in Fig.\ref{fig:eCOS-toy}

\begin{eqnarray}
\label{eq:toy-seq}
\begin{array}{ll}
A = [a b a b a b a b] \\
B = [a a a a b b b b]  \\
C = [b b b b a a a a] \\
D = [b b a a]  \\
\end{array}
\end{eqnarray}

In Fig.\ref{fig:eCOS-toy} the extreme left part of the curves, characterized with high $\nu$ values, corresponds, when applied on sequences having the same length and represented by vectors, to the cosine similarity constructed with the Euclidean inner product. The right part of the curves, characterized with very low $\nu$ values, corresponds to the cosine similarity evaluated using the \textit{tf(-idf)} vector space model. The center part of the figure, characterized with medium $\nu$ values shows that \textit{elasticity} allows to better discriminating between the sequences.

\section{Experimentations}
\label{Expe}
\subsection{Time series classification}
We empirically evaluate the effectiveness of the distance induced by an elastic inner product comparatively to the Euclidean and the Dynamic Time Warping distances using some classification tasks on a set of time series coming from quite different application fields.  The classification task we have considered consists of assigning one of the possible categories to an unknown time series for the 20 data sets available at the UCR repository \cite{TWED:KeoghUCRdataset}. As time is not explicitly given for these datasets, we used the index value of the samples as the timestamps for the whole experiment.\\

For each dataset, a training subset (TRAIN) is defined as well as an independent testing subset (TEST). We use the training sets to train two kinds of classifiers:
\begin{itemize}
 \item the first one is a first near neighbor (1-NN) classifier: first we select a training data set containing time series for which the correct category is known. To assign a category to an unknown time series selected from a testing data set (different from the train set), we select its nearest neighbor (in the sense of a distance or similarity measure) within the training data set, then, assign the associated category to its nearest neighbor. For that experiment, a leave one out procedure is performed on the training dataset to optimize the meta parameter $\nu$ of the considered elastic distance.
 \item the second one is a SVM classifier \cite{TWED:BoserGV92}, \cite{TWED:Vapnik1995} configured with a Gaussian RBF kernel whose parameters are $C > 0$, a trade-off between regularization and constraint violation and $\sigma$ that determines the width of the Gaussian function. To determine the $C$ and $\sigma$ hyper parameter values, we adopt a 5-folded cross-validation method on each training subset. According to this procedure, given a predefined training set TRAIN and a test set TEST, we adapt the meta parameters based on the training set TRAIN: we first divide TRAIN into 5 stratified subsets ${TRAIN_1, TRAIN_2,\cdots, TRAIN_5}$; then for each subset $TRAIN_i$ we use it as a new test set, and regard $(TRAIN-TRAIN_i)$ as a new training set; Based on the average error rate obtained on the five classification tasks, the optimal values of meta parameters are selected as the ones leading to the minimal average error rate. For the elastic kernel, the meta parameter $\nu$ is also optimized using this 5-folded cross-validation method performed on the training datasets. \\
\end{itemize}

The classification error rates are then estimated on the TEST datasets on the basis of the parameter values optimized on the TRAIN datasets.
We have used the LIBSVM library \cite{TWED:Libsvm01} to implement the SVM classifiers.

We tested the time elastic inner product $<A, B>_{eip}$ (Eq.\ref{Eq_eip}).  Precisely, we used the time-warp distance induced by  $<A, B>_{eip}$, basically $\delta_{eip}(A, B)=\left(<A-B, A-B>_{eip}\right)^{1/2}=(<A, A>_{eip}+<B, B>_{eip} - 2.<A, B>_{eip})^{1/2}$.\\ 

To speed up the computation at exploitation stage, we have used the construction proposed in Sec.\ref{IR} Prop.\ref{PropRnEmbedding} with the following \textit{elastic} Matrix $E$, where $\nu>0$ :

\begin{flalign*}
& E = \\
&\left[ {\begin{array}{ccccc}
 1.0 & e^{-\nu|t_{1}-t_{2}|^{2}} & e^{-\nu|t_{1}-t_{3}|^{2}} &  \cdots & e^{-\nu|t_{1}-t_{n}|^{2}}\\
 e^{-\nu|t_{2}-t_{1}|^{2}} & 1.0 & e^{-\nu|t_{2}-t_{3}|^{2}} &  \cdots & e^{-\nu|t_{2}-t_{n}|^{2}} \\
 e^{-\nu|t_{3}-t_{1}|^{2}} &e^{-\nu|t_{3}-t_{2}|^{2}} & 1.0 &  \cdots & e^{-\nu|t_{3}-t_{n}|^{2}} \\
 \cdots &  \cdots & \cdots & \cdots & \cdots  \\
 e^{-\nu|t_{n}-t_{1}|^{2}} &e^{-\nu|t_{n}-t_{2}|^{2}} &  e^{-\nu|t_{n}-t_{3}|^{2}} &  \cdots & 1.0 \\
\end{array} } \right]
\end{flalign*}

The databases $D_E=\{E B(i)_{1}^{n}\}_{i=1 \cdots m}$ are thus constructed \textit{off-line} from the TRAIN datasets, once the optimization procedure of the $\nu$ parameter has been completed.  $D_E$ is then exploited \textit{on-line} by the 1-NN and SVM classifiers.

\begin{table*}[!ht]
 \caption{Dataset sizes and meta parameters used in conjunction with  $K_{ed}$, ${K}_{eip}$ and ${K}_{dtw}$  kernels}
\label{MetaOptiVal}
\centering
\begin{tabular}{|l|c|c|c|c|}
\hline
\textbf{DATASET} & \textbf{length$|$\#class$|$\#train$|$\#test} &  \textbf{$K_{ed}: C, \sigma$}  &  \textbf{$K_{Deip}: \nu, C, \sigma$} &  \textbf{$K_{DTW}: C, \sigma$} \\
\hline
\textbf{Synthetic control} & 60$|$6$|$300$|$300 & 1.0;0.125 & .1;.25;.0625 &  8.0;4.0\\
\hline
\textbf{Gun-Point} & 150$|$2$|$50$|$150 & 256;.5 & 0.01;256;.5  & 16.0;0.0312\\
\hline
\textbf{CBF} & 128$|$3$|$30$|$900 & 8;1.0 & .001;4.0;.0312 & 1.0;1.0\\
\hline
\textbf{Face (all)} & 131$|$14$|$560$|$1690 & 4;0.5 & .1;8.0;.5  & 2.0;0.25\\
\hline
\textbf{OSU Leaf} & 427$|$6$|$200$|$242 & 2;0.125 & .1;4;0.125 & 4.0;0.062\\
\hline
\textbf{Swedish Leaf} & 128$|$15$|$500$|$625 & 128;0.125 & 10;8;.0625 & 4.0;0.031\\
\hline
\textbf{50 Words} & 270$|$50$|$450$|$455 & 32;0.5 & 0.01;32;0.5 & 4.0;0.062\\
\hline
\textbf{Trace} & 275$|$4$|$100$|$100 & 8;0.0156 & .001;256;.0625 & 4;0.25\\
\hline
\textbf{Two Patterns} & 128 $|$4$|$1000$|$4000 & 4.0;0.25 & .01;1;.0312 & 0.25,0.125\\
\hline
\textbf{Wafer} & 152$|$2$|$1000$|$6174 & 4.0;0.5 & 0.01;32;.5 & 1.0;0.016\\
\hline
\textbf{face (four)} & 350$|$4$|$24$|$88 & 8.0;2.0 & .01;16;2 & 16;0.5\\
\hline
\textbf{Ligthing2} & 637$|$2$|$60$|$61 & 2.0;0.125 & .001;128;2  & 2.0;0.031\\
\hline
\textbf{Ligthing7} & 319$|$7$|$70$|$73 & 32.0;256.0 & .1;16;.5  & 4;0.25\\
\hline
\textbf{ECG200} & 96$|$2$|$100$|$100 & 8.0;0.25 & 0, 1024, .0625  & 2;0.62\\
\hline
\textbf{Adiac} & 176$|$37$|$390$|$391 & 1024.0;0.125 & 10;256.0;.0312  & 16;0.0039\\
\hline
\textbf{Yoga} & 426$|$2$|$300$|$3000 & 64.0;0.125 & 1;32;.0625  & 4;0.008\\
\hline
\textbf{Fish} & 463$|$7$|$175$|$175 & 64.0;1.0 & .01;256;1  & 8;0.016\\
\hline
\textbf{Coffee} & 286$|$2$|$28$|$28 & 128.0;4.0 & .01;1024;4 & 8;0.062\\
\hline
\textbf{OliveOil} & 570$|$4$|$30$|$30 & 2.0;0.125 & .01;64;2  & 2;0.125\\
\hline
\textbf{Beef} & 470$|$5$|$30$|$30 & 128.0;4.0 & 0;64;.5 & 16;0.016\\
\hline
\end{tabular}
\end{table*}

\subsubsection{Meta parameters}

$\delta_{eip}$ is characterized by the meta parameter $\nu$ (the \textit{stiffness} parameter) that is optimized for each dataset on the train data by minimizing the classification error rate of a first near neighbor classifier. For this kernel, $\nu$ is selected in $\{100, 10, 1, .1, .01, ..., 1e-5, 0\}$. \\

To explore the potential benefits of an elastic inner product against the Euclidean inner product,  we also tested the Euclidean Distance $\delta_{ed}$ which, as already stated, is the limit when $\nu \rightarrow \infty$ of $\delta_{eip}$.\\

The kernels exploited by the SVM classifiers are the Gaussian kernels $K_{eip}(A,B)=e^{-\delta_{eip}(A,B)^2/(2\cdot\sigma^2)}$ and $K_{ed}(A,B)=e^{-\delta_{ed}(A,B)^2/(2\cdot\sigma^2)}$. The meta parameter $C$ is selected from the discrete set $\{2^{-8}, 2^{-7}, ..., 1,2, ..., 2^{10}\}$, and $\sigma^2$ from $\{2^{-5}, 2^{-4}, ..., 1,2, ..., 2^{10}\}$. 

Table \ref{MetaOptiVal} gives for each data set and each tested kernels ($K_{ed}$ and $K_{eip}$) the corresponding optimized values of the meta parameters. \\

\begin{table*}[!ht]
 \caption{Comparative study using the UCR datasets: classification error rates (in \%) obtained using the first near neighbor (1-NN) classification rule  and a SVM classifier for the $K_{ed}$,  ${K}_{eip}$ and ${K}_{eip}$ kernels. Two scores are given S1$|$S2: the first one, S1, is evaluated on the training data, while the second one, S2, is evaluated on the test data. For the each classification methods (1-NN and SVM) the rank of each classifier is given in parenthesis ((1): best classifier, (2): 2nd best classifier, (3): 3rd best classifier}
\label{TabRes}
\centering
\begin{tabular}{|l|c|c|c||c|c|c|c|c|}
\hline
\textbf{DATASET} & 1-NN \textbf{$\delta_{ed}$} & \textbf{1-NN $\delta_{eip}$} & \textbf{1-NN $\delta_{dtw}$} & \textbf{SVM $K_{ed}$} & \textbf{SVM $K_{eip}$} & \textbf{SVM $K_{dtw}$} \\
\hline\hline
\textbf{Synthetic control} & 9(3)$|$12(3) & .67(1)$|$1(2) & 1.0(2)$|$0.67(1) & 3(3)$|$2.33(3) & .33(2)$|$.67(1) & 0(1)$|$1(2)\\
\hline
\textbf{Gun-Point} & 4.0(1)$|$8.67(1)  & 4.0(1)$|$8.67(1) & 18.36(3)$|$9.3(3) & 4.0(3)$|$6.0(3) & 2.0(2)$|$2.67(2) & 0(1)$|$1.33(1)\\
\hline
\textbf{CBF} & 16.67(3)$|$14.78(3) & 3.33(2)$|$4.22(2) & 0(1)$|$0.33(1) & 3.33(2)$|$10.89(3) & 3.33(1)$|$5(1) & 3.33(1)$|$5.44(2)\\
\hline
\textbf{Face (all)} & 11.25(3)$|$28.64(3) & 7.5(2)$|$26.33(2) & 6.8(1)$|$19.23(1) & 9.82(3)$|$16.45(3) & 6.25(2)$|$24.91(2) & .54(1)$|$16.98(1)\\
\hline
\textbf{OSU Leaf} & 37.0(2)$|$48.35(2) & 37(2)$|$48.35(2) & 33.17(1)$|$40.9(1) & 35(3)$|$44.21(2) & 34.5(2)$|$44.21(2) & 20(1)$|$23.55(1)\\
\hline
\textbf{Swedish Leaf} & 26.6(3)$|$21.12(3) & 24.4(1)$|$20.96(2) & 24.65(2)$|$20.8(1) & 15(2)$|$8.64(2) & 15(2)$|$8.64(2) & 7(1)$|$5.6(1)\\
\hline
\textbf{50 Words} & 34.47(3)$|$36.32(3) & 32.2(1)$|$32.73(2) & 33.18(2)$|$31(1) & 33.56(3)$|$30.99(3) &  31.78(2)$|$29.67(2) & 15.21(1)$|$17.58(1)\\
\hline
\textbf{Trace} & 18(3)$|$24(2) & 16(2)$|$24(2) & 0(1)$|$0(1) & 9(3)$|$19(3) & 1(2)$|$7(2) & 0(1)$|$2(1)\\
\hline
\textbf{Two Patterns} & 8.5(3)$|$9.32(3) & 4.3(2) $|$3.62(2) & 0(1)$|$0(1) & 8.6(3)$|$7.45(3) & 5.5(2)$|$3.52(2) & 0(1)$|$0(1)\\
\hline
\textbf{Wafer} & 0.7(2)$|$0.45(2) & .5(1)$|$.42(1) & 1.4(3)$|$2.01(3) & .7(3)$|$.7(3) & .2(2)$|$.68(2) & 0(1)$|$0.39(1)\\
\hline
\textbf{face (four)} & 37.5(3)$|$21.59(3) & 33.33(2)$|$19.31(2) &26.09(1)$|$17.05(1) & 20.84(3)$|$19.31(3) & 16.67(2)$|$13.63(2) & 8.33(1)$|$5.68(1)\\
\hline
\textbf{Ligthing2} & 25.0(3)$|$24.59(3) & 20(2)$|$16.39(2) & 13.56(1)$|$13.1(1) & 21.77(3)$|$31.14(3) & 20(2)$|$26.22(2) & 8.33(1)$|$19.67(1)\\
\hline
\textbf{Ligthing7} & 35.71(3)$|$42.47(3) & 30.0(1)$|$32.87(2) & 33.33(2)$|$27.4(1) & 37.14(3)$|$36.98(3)  & 34.29(2)$|$35.61(2) & 17.14(1)$|$16.43(1) \\
\hline
\textbf{ECG200} &14.0(2)$|$12.0(2) &  1.0(1)$|$2.0(1) & 23.23(3)$|$23(3) & 8.0(3)$|$9.0(2)  & 3.0(1)$|$7.0(1) & 7(2)$|$13(3)\\
\hline
\textbf{Adiac} & 41.28(3)$|$38.87(1) & 39.59(1)$|$38.87(1) & 40.62(2)$|$39.64(3) & 26.67(3)$|$24.04(1) & 25.13(2)$|$24.04(1)  & 24.61(1)$|$25.32(3) \\
\hline
\textbf{Yoga} & 22.67(3)$|$16.9(2) & 21.67(2)$|$22.26(3) & 16.37(1)$|$16.4(1) & 17.66(3)$|$14.43(2) & 15.33(2)$|$14.4(2) & 11(1)$|$11.2(1)\\
\hline
\textbf{Fish} & 24.0(1)$|$21.71(2) & 24.0(1)$|$21.71(2) & 26.44(3)$|$16.57(1) & 14.86(3)$|$13.14(3) & 13.14(2)$|$12.57(2) & 6.86(1)$|$4.57(1)\\
\hline
\textbf{Coffee} & 21.43(2)$|$25.0(2) & 21.43(2)$|$25.0(2) & 14.81(1)$|$17.86(1) & 0(1)$|$0(1) & 0(1)$|$7.14(2) & 10.71(3)$|$17.86(3)\\
\hline
\textbf{OliveOil} & 13.33(1)$|$13.33(1) & 13.33(1)$|$13.33(1) & 13.79(3)$|$13.33(1) & 10.0(1)$|$10.0(1) & 10.0(1)$|$10.0(1) & 13.33(3)$|$16.67(3)\\
\hline
\textbf{Beef} & 46.67(1)$|$46.67(1) & 46.67(1)$|$46.67(1) & 55.17(3)$|$50(3) & 37.67(2)$|$30(1) & 37.67(2)$|$30(1) & 32.14(1)$|$42.85(3)\\
\hline
\hline
\textbf{Average Rank} & (2.4)$|$(2.25) & (1.45)$|$(1.75) & (1.85)$|$(1.5) & (2.65)$|$(2.4) & (1.8)$|$(1.7) & (1.25)$|$(1.6)\\
\hline
\end{tabular}
\end{table*}

\begin{figure*}[!ht]
\centering
\includegraphics[scale=1.]{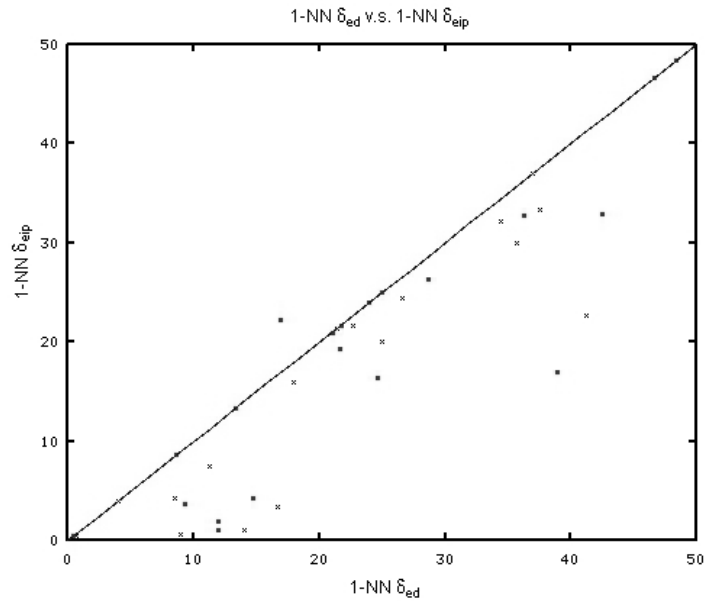}
\caption{Comparison of error rates (in \%) between two 1-NN classifiers based on the Euclidean Distance (1-NN $\delta_{ed}$), and the distance induced by a time-warp inner product (1-NN  $\delta_{eip}$). The straight line has a slope of 1.0 and dots correspond, for the pair of classifiers, to the error rates on the train (star) or test (circle) data sets. A dot below (resp. above) the straight line indicates that distance $\delta_{eip}$ has a lower (resp. higher) error rate than distance $\delta_{ed}$. Black 'x' indicate error rates evaluated on the training data, while the red squares indicate error rates evaluated on the test data.}
\label{fig:1NNEDvsDEIP}
\end{figure*}

\begin{figure*}[!ht]
\centering
\includegraphics[scale=1.]{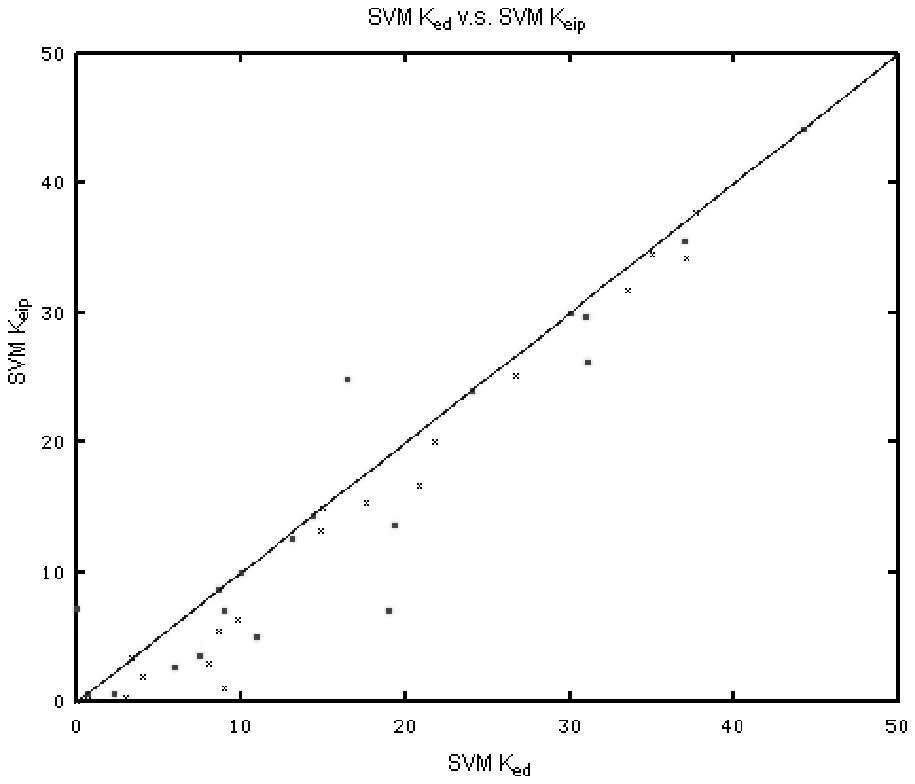}
\caption{Comparison of error rates (in \%) between two SVM classifiers, the first one based on the Euclidean Distance Gaussian kernel (SVM $K_{ed}$), and the second one based on a Gaussian kernel induced by a time-warp inner product (SVM $K_{eip}$). The straight line has a slope of 1.0 and dots correspond, for the pair of classifiers, to the error rates on the train (star) or test (circle) data sets. A dot below (resp. above) the straight line indicates that SVM $K_{eip}$ has a lower (resp. higher) error rate than distance SVM $K_{ed}$. Black 'x' indicate error rates evaluated on the training data, while the red squares indicate error rates evaluated on the test data.}
\label{fig:SVMEDvsDEIP}
\end{figure*} 

According to the classification results, this experiment shows that the distance induced by the elastic inner product $\delta_{eip}$ is significantly more effective for the considered tasks comparatively to the Euclidean distance. It exhibits, on average, the lowest error rates for most of the tested datasets and for both the 1-NN and SVM classifiers, as shown in Table \ref{TabRes} and Figures \ref{fig:1NNEDvsDEIP} and \ref{fig:SVMEDvsDEIP}. The stiffness parameter $\nu$ in $\delta_{eip}$ seems to play a significant role in these classification tasks, and this for a quite large majority of data sets.\\

Only one dataset, \textit{yoga}, is better classified by the 1-NN $\delta_{ed}$ classifier on the test data, although the error rate on the train data is lower for the 1-NN $\delta_{eip}$ classifier. For the SVM classifiers, only two datasets, \textit{Face (all)} and \textit{Coffee}, are significantly better classified on the test data by SVM $K_{ed}$ classifiers. Nevertheless, for these two datasets, $K_{eip}$ reaches a better (or best) score on the train data. We are facing here the trade-off between learning  and generalization capabilities. The meta parameter $\nu$ is selected such as to minimized the classification error on the train data. If this strategy is on average a winning strategy, some datasets show that it does not always lead to a good trade-off, this is the case for \textit{Face (all)} and \textit{yoga} datasets. 

However, $\delta_{dtw}$ based classifiers outperform $\delta_{eip}$ based classifiers on a majority of datasets. The average ranking of the classifiers shows clearly that $\delta_{eip}$ ranks in between $\delta_{ed}$ and $\delta_{dtw}$. $\delta_{eip}$ can be considered as a compromise between the linear Euclidean distance and the non-linear $\delta_{dtw}$. It maintains a low computational cost and nevertheless compensates some of the limitations of the Euclidean distance that it is very sensitive to distortions in the time axis. It should be noted that $\delta_{eip}$ can cope with sample substitution, deletion and insertion as well as $\delta_{dtw}$. In addition, $\delta_{eip}$ can deal with sample permutations while $\delta_{dtw}$ cannot.

\begin{table*}[!ht]
 \caption{Comparative study using the PCBC protein datasets available at http://hydra.icgeb.trieste.it/benchmark. The assessment measure is the average AUC values based on the ROC curves obtained by the first near neighbor classification rules when BLAST, SW, NW, LA, PRIDE and eCOS($\nu$=1,.1,.05, .01, .001) similarity measures are used. Best average AUC values are in boldface characters.}
\label{TabSeqClass}
\centering
\begin{tabular}{|l|c|c|c|c|c||c|c|c|c|c|c|}
\hline
\textbf{1NN}  &	\textbf{BLAST}  &	\textbf{SW}	&  \textbf{NW}  & 	\textbf{LA} &	\textbf{PRIDE}	& \textbf{$(\nu=1)$} & \textbf{$(\nu=.1)$} & \textbf{$(\nu=.05)$} & \textbf{$(\nu=.01)$} & \textbf{$(\nu=.001)$} & \textbf{$(Best)$} \\
\hline\hline
SCOP95\_supfam\_fam\_1	&	70.01	&	83.52	&	83.95	&	78.78	&	\textbf{89.48}	&	77.70	&	81.99	&	82.94	&	83.63	&	80.74	&	83.63 	\\
SCOP95\_supfam\_fam\_kfold\_2	&	94.87	&	\textbf{97.74}	&	97.73	&	96.71	&	97.41	&	91.02	&	94.84	&	95.02	&	94.15	&	91.44	&	95.02 	\\
SCOP95\_fold\_supfam\_3	&	56.87	&	64.82	&	67.47	&	58.38	&	\textbf{84.72}	&	75.20	&	79.27	&	79.09	&	75.71	&	69.89	&	79.27 	\\
SCOP95\_fold\_supfam\_kfold\_4	&	90.30	&	93.77	&	94.54	&	91.38	&	\textbf{95.82}	&	88.93	&	93.40	&	93.32	&	90.67	&	85.79	&	93.4 	\\
SCOP95\_class\_fold\_5	&	60.18	&	61.04	&	62.90	&	56.90	&	\textbf{78.27}	&	59.25	&	65.65	&	67.13	&	67.23	&	64.45	&	67.23 	\\
SCOP95\_class\_fold\_kfold\_6	&	92.23	&	92.14	&	92.85	&	88.28	&	\textbf{93.24}	&	80.43	&	88.38	&	88.76	&	85.32	&	79.81	&	88.76 	\\
CATH95\_H\_S\_7	&	97.90	&	99.20	&	\textbf{99.29}	&	98.81	&	95.47	&	95.42	&	97.03	&	96.84	&	95.78	&	91.81	&	97.03 	\\
CATH95\_H\_S\_kfold\_8	&	96.63	&	98.30	&	\textbf{98.32}	&	97.80	&	95.02	&	95.10	&	97.28	&	97.22	&	96.36	&	93.67	&	97.28 	\\
CATH95\_T\_H\_9	&	57.33	&	64.75	&	66.97	&	61.44	&	76.71	&	73.25	&	78.57	&	\textbf{79.12}	&	77.38	&	70.20	&	79.12 	\\
CATH95\_T\_H\_kfold\_10	&	89.76	&	92.60	&	\textbf{93.14}	&	90.66	&	90.96	&	88.98	&	93.02	&	92.78	&	90.28	&	84.25	&	93.02 	\\
CATH95\_A\_T\_11	&	54.52	&	56.82	&	57.97	&	54.12	&	\textbf{67.37}	&	57.78	&	61.63	&	62.09	&	61.33	&	59.70	&	62.09 	\\
CATH95\_A\_T\_kfold\_12	&	90.21	&	91.28	&	\textbf{91.77}	&	88.91	&	84.89	&	82.44	&	88.28	&	88.14	&	84.96	&	80.09	&	88.28 	\\
CATH95\_C\_A\_13	&	64.81	&	\textbf{71.28}	&	69.87	&	69.17	&	55.72	&	37.74	&	46.33	&	49.79	&	62.61	&	63.63	&	63.63 	\\
CATH95\_C\_A\_kfold\_14	&	90.80	&	89.74	&	\textbf{90.51}	&	86.21	&	80.63	&	76.75	&	83.29	&	83.94	&	80.57	&	75.71	&	83.94 	\\
\hline
\textbf{Average}	&	79.03	&	82.64	&	83.38	&	79.83	&	\textbf{84.69}	&	77.14	&	82.07	&	82.58	&	81.86	&	77.94	&	84.00	\\
\hline
\end{tabular}
\end{table*}

\subsection{Sequence classification}
We report here a protein classification experiment carried out using the Protein Classification Benchmark Collection (PCBC) \cite{TWED:PCBC2007} \cite{TWED:PCBC-URL}. This benchmark contains structural and functional annotations of proteins. The two datasets that we have exploited, SCOP95 and CATH95, are available at http://hydra.icgeb.trieste.it/benchmark.

The entries of the SCOP95 dataset are characterized by sequences with variable lengths and relatively little sequence similarity (less than 95\% sequence identity) between the protein families. 

The CATH95 dataset contains near-identical protein families of variable lengths in which the proteins have a high sequence similarity (more than 95\% sequence identity). 
 
Basically, the considered classification tasks involve protein domain sequence and structure comparisons at various levels of the structural hierarchies. We have considered the following 14 PCB subsets:
\begin{itemize}
	 \item PCB00001 	$SCOP95\_Superfamily\_Family $	
	 \item PCB00002 	$SCOP95\_Superfamily\_5fold$ 	
	 \item PCB00003 	$SCOP95\_Fold\_Superfamily$ 	
	 \item PCB00004 	$SCOP95\_Fold\_5fold$ 	
	 \item PCB00005 	$SCOP95\_Class\_Fold$ 	
	 \item PCB00006 	$SCOP95\_Class\_5fold$ 	
	 \item PCB00007 	$CATH95\_Homology\_Similarity$ 	
	 \item PCB00008 	$CATH95\_Homology\_5fold$ 	
	 \item PCB00009 	$CATH95\_Topology\_Homology$ 	
	 \item PCB00010 	$CATH95\_Topology\_5fold$ 	
	 \item PCB00011 	$CATH95_Architecture\_Topology$ 	
	 \item PCB00012 	$CATH95\_Architecture\_5fold$ 	
	 \item PCB00013 	$CATH95\_Class\_Architecture$ 	
	 \item PCB00014 	$CATH95\_Class\_5fold$
\end{itemize}

We evaluate the elastic cosine similarity based on the $eip$ defined for symbolic sequences (Def.\ref{defEIP_TM}, Eq.\ref{eqEIP_TM}) comparatively to five other similarity measures commonly used in Bioinformatics:
\begin{itemize}
    \item BLAST \cite{TWED:BLAST1990}: the Basic Local Alignment Search Tool is a very popular family of fast heuristic search methods used for finding similar regions between two or more nucleotides or amino acids.

    \item SW \cite{TWED:SmithWaterman81}: The Smith–Waterman algorithm is used for performing local sequence alignment, for determining similar regions between two nucleotide or protein sequences. Instead of looking at the sequence globally as NW does, the Smith–Waterman algorithm compares subsequences of all possible lengths.
    
    \item NW \cite{TWED:NeedlemanWunschNW70}: The Needleman Wunsch algorithm performs a maximal global alignment of two strings. It is commonly used in bioinformatics to align protein sequences or nucleotides.     
    
    \item LA kernel \cite{TWED:Saigo04-LA}: The Local Alignment kernel is used to detect local alignment between strings by convolving simple basic kernels. Its construction mimic the local alignment scoring schemes proposed in the SW algorithm.
        
    \item PRIDE \cite{TWED:Carugo98-PRIDE}: The PRIDE score is estimated as the PRobability of IDEntity between two protein 3D structures. The calculation of similarity between two proteins, is based on the comparison of histograms of the pairwise distances between $C-\alpha$ residues whose distribution is highly characteristic of protein folds.
    
\end{itemize}

The average AUC (area under the ROC Curve) measure is evaluated for 1-NN classifiers exploiting respectively BLAST, SW, NW, LA, PRIDE and $eCOS(\nu)$  as alignment methods. 
One can notice that these datasets are quite well suited for global alignment since, as shown in table \ref{TabSeqClass}, the NW algorithm performs better than the SW algorithm. The $eip$ structure that considers global alignment is thus well adapted to the task.
We show on these experiments that for $\nu=.05$ the $eCOS$ classifier in average performs significantly better than BLAST heuristics \cite{TWED:BLAST1990} and LA \cite{TWED:Saigo04-LA}, the local alignment kernel for string. Furthermore, it performs almost as well as the SW and NW algorithms. The PRIDE method \cite{TWED:Carugo98-PRIDE} gets the best results, but it uses the tertiary structure of the proteins while all the other methods exploit the primary structure. Here again, a ranking based on $eCOS$ similarity has a complexity that could be maintained linear at exploitation stage (i.e. when testing numerous sequences against massive datasets). These very positive results offer perspective in fast filtering of biological symbolic sequences.

\subsection{Experimental complexity}
\begin{figure*}[ht!]
\centering
\includegraphics[scale=0.45]{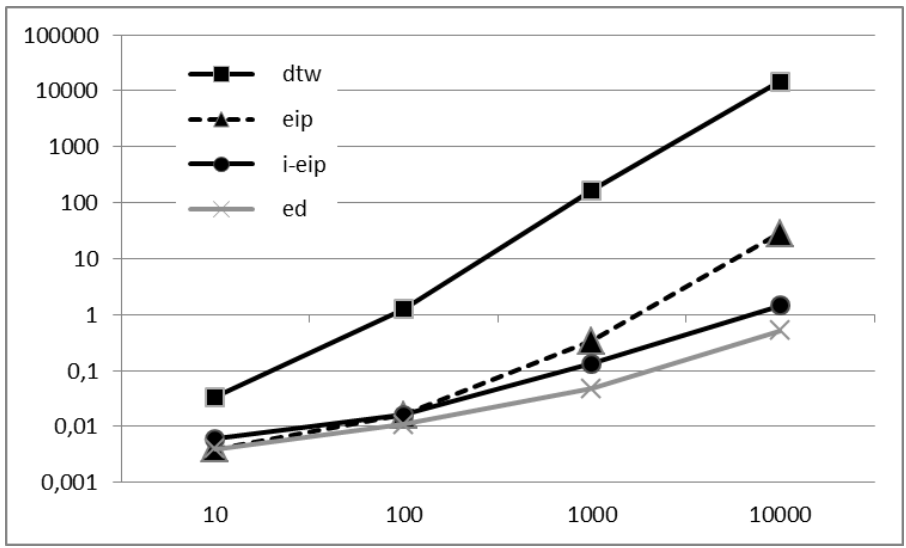}
\caption{Elapsed time in second as a function of the time series length for 10000 distance computations and for the four experimented distances: $\delta_{ed}$ (cross symbol, plain grey line), $\delta_{eip}$ (triangle symbol plain line), $\delta_{i-eip}$ (round symbol, dashed lines) and $\delta_{dtw}$(square symbol, plain line).} 
\label{fig:perfs}
\end{figure*}

To evaluate in practice the computational cost of $\delta_{eip}$, we compare it with two other distances, namely $\delta_{ed}$ (the Euclidean distance) which has a linear complexity, and $\delta_{dtw}$, the Dynamic Time Warping distance which has a quadratic complexity. In addition we evaluate the computational cost of the indexed version of $\delta_{eip}$ that we refer to as $\delta_{i-eip}$. The experiment consists in producing random datasets of 100 time series of increasing lengths (10, 100, 1000 and 10000 samples) and computing the 100x100 distance matrices. Figure \ref{fig:perfs} gives the elapsed time in second, according to a logarithmic scale, for the four distances as a function of the length of the time series. When compared to $\delta_{dtw}$, $\delta_{eip}$ is evaluated very efficiently using the matrix computation given in Eq.\ref{Eq.matrixEIP}, although, without any off line indexing of the time series, $\delta_{eip}$ cannot compete with $\delta_{ed}$ when the length of the time series increases. The $\delta_{i-eip}$ curve has clearly the same slope than the $\delta_{ed}$ curve, showing that the off-line indexed version of $\delta_{eip}$ is characterized with a linear complexity, that includes a nevertheless linear overhead when compared to $\delta_{ed}$, mainly due to the loading of the index.

\section{Conclusion}

 This paper has proposed what we call a family of \textit{elastic inner products} able to cope with non-uniformly sampled time series of various lengths, as far as they do not contain the \textit{zero} or $null$ symbol value. These constructions allow one to embed any such time series in a single inner space, that some how generalizes the notion of Euclidean inner space. The recursive structure of the proposed construction offers the possibility to manage several \textit{elastic} dimensions. Some applicative benefits can be expected in time series or sequence analysis when time \textit{elasticity} is an issue, for instance in the field of numeric or symbolic sequence data mining.

 If the algorithmic complexity required to evaluate an elastic inner product is in general quadratic, its computation can be much more efficiently performed than dynamic programming algorithms. In addition we have shown that for some information retrieval applications for which embeddings of time series or symbolic sequences into a finite dimensional Euclidean space is possible,  one can break this quadratic complexity down to a linear complexity at exploitation time, although a quadratic computational cost should still be paid once during a preprocessing step at indexing phase.
 
 The preliminary experiments we have carried out on some time series and symbolic sequence classification tasks show that embedding time series into elastic inner product space may brought significant accuracy improvement when compared to Euclidean inner product space embeddings as they compensate, at least partially, the limitations of Euclidean distance which is very sensitive to distortions in the time axis. Although our experiments show that Dynamic Programming matching algorithms outperforms on a majority of dataset distances that are derived from an elastic inner product, on some datasets such distances lead to similar accuracies .

 The experiment carried out on symbolic sequences alignment involves sequences of various lengths. It shows also some very interesting perspectives in the scope of fast filtering of massive data, since the accuracy obtained by a 1-NN symbolic \textit{elastic cosine} classifier with a potentially linear complexity at exploitation time is somehow comparable to the one obtained using dynamic programming algorithms (NW and SW) whose complexity are quadratic when the alignment search space is not restricted.

 Finally, the general recursive structure of the elastic inner product opens perspectives in the processing of more complex data such as tree data mining for which considering several elastic dimensions may be relevant and efficient. The possibility to decompose complex structures onto sets of elastic basis vectors opens perspectives in various areas of application such as data compression, multi-dimensional scaling or matching pursuits.




\bibliography{EVS}{}
\bibliographystyle{plain}

\end{document}